# Who Is the Agent Talking To? Participant-Aware Role Switching for Embodied Student-Support Dialogue

Iro Tochukwu Samuel
Edinburgh Napier University
Edinburgh, United Kingdom
40771994@live.napier.ac.uk

Filip Janik
Edinburgh Napier University
Edinburgh, United Kingdom
40534678@live.napier.ac.uk

Ewa Olton
Edinburgh Napier University
Edinburgh, United Kingdom
40679395@live.napier.ac.uk

Robert Smales
Edinburgh Napier University
Edinburgh, United Kingdom
40596040@live.napier.ac.uk

Harris Spratt
Edinburgh Napier University
Edinburgh, United Kingdom
40530696@live.napier.ac.uk

Shea Tait
Edinburgh Napier University
Edinburgh, United Kingdom
40535201@live.napier.ac.uk

Md Zia Ullah
Edinburgh Napier University
Edinburgh, United Kingdom
m.ullah@napier.ac.uk

Yanchao Yu
Edinburgh Napier University
Edinburgh, United Kingdom
y.yu@napier.ac.uk

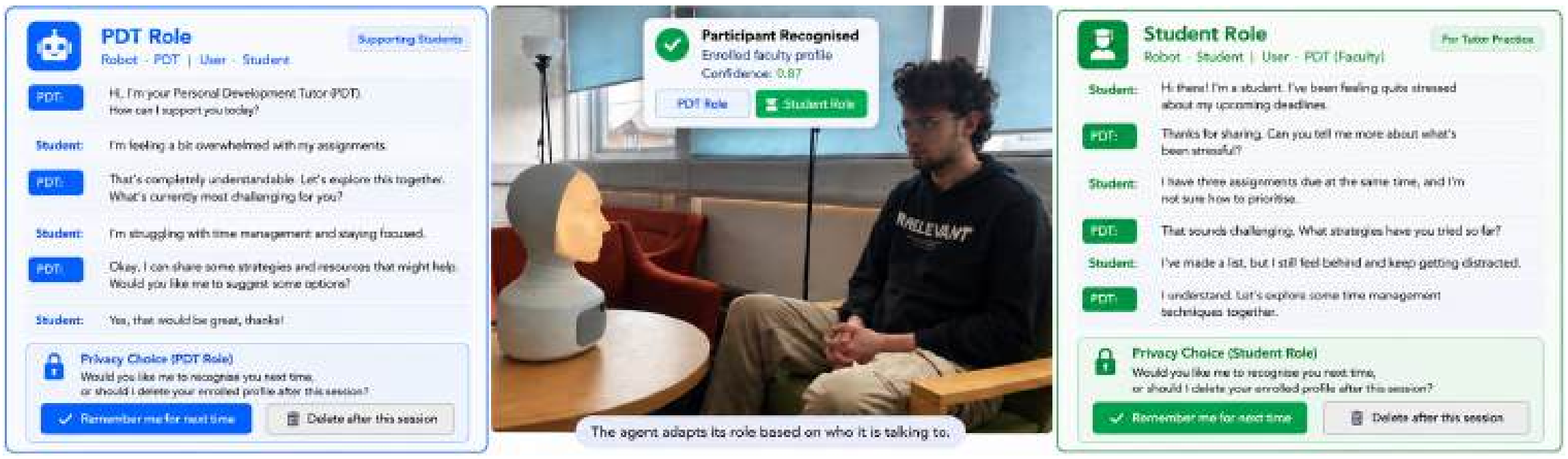

**Figure 1: RPO-PDT showing PDT support, faculty role-play, participant recognition, and conversational privacy control.**

## Abstract

We present RPO-PDT, an embodied conversational agent that adapts its conversational role to the current participant. It acts as a Personal Development Tutor (PDT) when supporting students and as a simulated student when faculty rehearse tutoring strategies. Voluntary closed-set participant verification informs a confidence-controlled role manager: stable enrolled-faculty matches activate the Student role, while enrolled students and unmatched participants use the default PDT role and uncertain observations cause no transition. The system combines Furhat interaction, Rasa orchestration, institutional retrieval, safeguarding, and LLM-based generation. User control is preserved through manual role override, consent-based facial enrolment, temporary interaction identifiers, and explicit retention or deletion of enrolled profiles. The demonstration exposes both role-switching behaviour and these privacy controls in interaction.

## 1 Introduction

Embodied conversational agents increasingly combine speech, vision, memory, and external knowledge. Prior work has explored multimodal social dialogue [6], persistent memory [11], and role-playing LLM agents [8–10]. Yet the conversational role of such systems is commonly fixed before interaction or selected manually. This is restrictive in higher-education support, where students may need institutionally grounded guidance while faculty may benefit from rehearsing tutoring strategies with a simulated student.

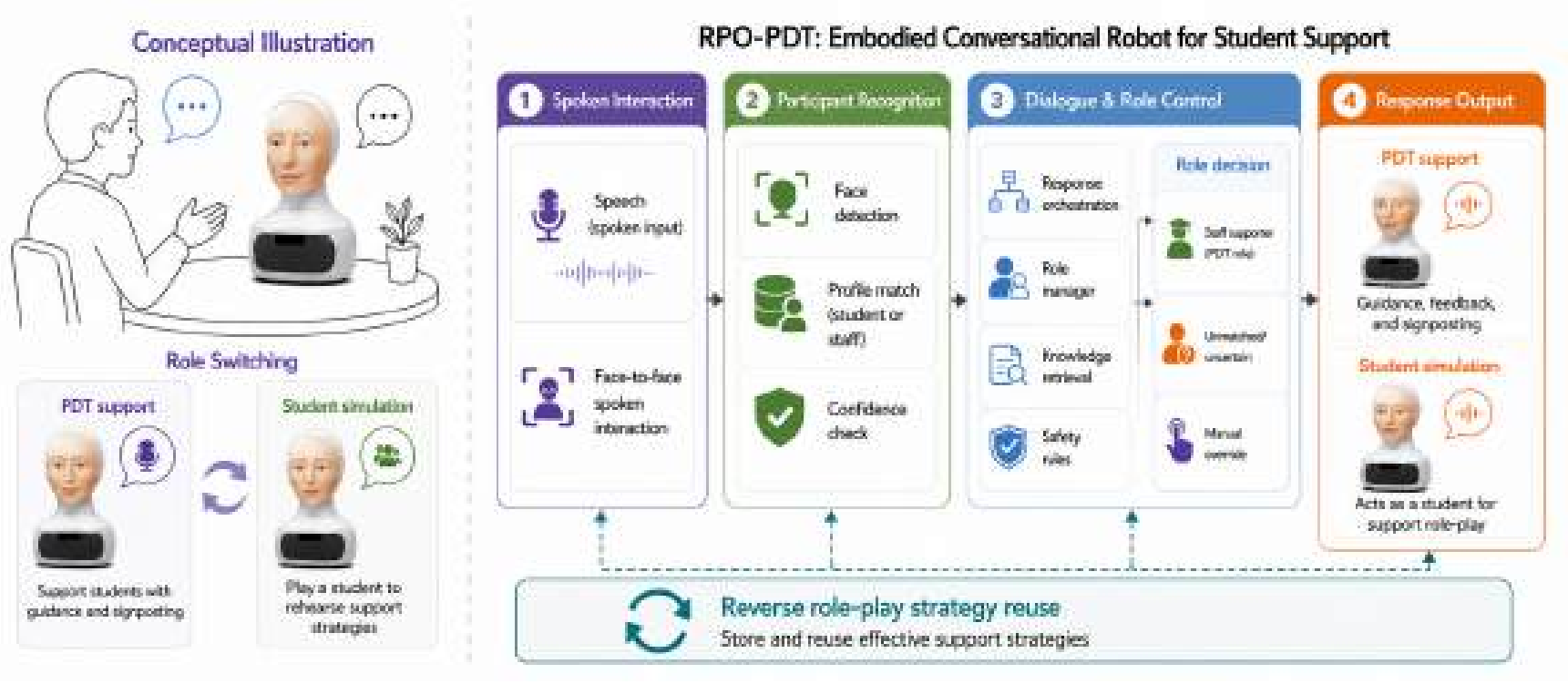

**Figure 2: Overview of the RPO-PDT architecture and interaction flow. Multimodal interaction, involving both audio and video, is combined with participant recognition to facilitate confidence-controlled role selection. When a participant is stably matched to an enrolled faculty (or student) profile, they activate the Student (or PDT) role. In contrast, unmatched or uncertain participants are assigned the default PDT role. The components of dialogue orchestration, institutional retrieval, safeguarding, response generation, and the reuse of reverse role-play strategies remain modular and open to inspection.**

We introduce RPO-PDT, a participant-aware embodied agent with two complementary roles. In the **PDT role**, the robot supports students; in the **Student role**, it simulates a student for faculty tutoring practice. Its central contribution is a confidence-controlled mechanism linking participant verification to explicit conversational-role management. A stable match to an enrolled faculty profile activates Student mode; an enrolled student or unmatched participant uses the default PDT role; and uncertain evidence preserves the current role. Manual role selection remains available throughout. Raw visual data are not passed to the LLM; instead, perception is reduced to a restricted dialogue-state signal that separates participant verification, role control, institutional grounding, safeguarding, and LLM-based response generation.

## 2 System and Interaction Design

Figure 2 shows the perception-to-dialogue pipeline. Furhat [1] provides face-to-face multimodal interaction, while Rasa [2] manages dialogue state and coordinates participant verification, institutional retrieval, safeguarding, and LLM (DeepSeek-V3.2 [7]) generation.

The vision component performs *closed-set participant verification*, not demographic or occupational classification. Participants voluntarily enrol reference images, which are converted to embeddings using a ResNet-50 [5] person re-identification model fine-tuned on VGGFace2 [4] and stored in a local SQLite gallery. During interaction, a Furhat-camera image is embedded and compared with enrolled representations using cosine similarity. The dialogue manager receives only the match state, confidence score, temporary participant identifier, role, and timestamp.

A role transition occurs only when the highest-scoring profile exceeds a similarity threshold of 0.40 and remains stable for three consecutive identification cycles (approximately 1.5 s). The threshold was selected using a disjoint VGGFace2 test set of 21,295 images from 60 unseen identities. The policy is defined as follows:

$$\text{Role} = \begin{cases} \text{Student}, & \text{stable enrolled-faculty match}, \\ \text{PDT}, & \text{stable enrolled-student or unmatched}, \\ \text{unchanged}, & \text{uncertain evidence}. \end{cases} \tag{1}$$

Manual commands can switch or restore the role, and the transition source is logged. In a functional test with three consented participants (two students and one faculty member), recognition confidence scores were 0.873, 0.822, and 0.673; all observed transitions were correct. Given the small sample, we treat this as a functional validation rather than a user study.

*Privacy and participant control.* Facial enrolment is voluntary and consent-based. Temporary participant identifiers are restricted to the current interaction, and unmatched participants are not automatically enrolled or assigned persistent identities. At the end of an enrolled participant's interaction, the robot can ask whether the recognition profile should be retained for future interactions or deleted, making persistence an explicit user decision. In the PDT role, institutional resource grounded responses through retrieval-augmented generation [3]; in the Student role, the agent adopts a stored student scenario. Potentially high-risk utterances are handled through deterministic escalation rules outside open-ended generation.

## 3 Demonstration Experience

The live demonstration exposes both the user-facing interaction and the underlying control logic (Figure 1). An unmatched or enrolled student interacts with the robot in the default PDT role, while recognition of an enrolled faculty participant triggers a stable transition to Student mode for tutoring practice. Attendees can also test uncertain recognition, such as partial views or confidence below the 0.40 threshold; in these cases, the current role is preserved. Manual override remains available, and participants can inspect the active match state, role, transition source, retrieved resource category, and safeguarding outcome.

Privacy control is demonstrated conversationally at the end of an enrolled participant's interaction, when the robot asks whether the recognition profile should be retained for future sessions or deleted. This choice is available in both PDT and Student interactions. The demonstration can also replay an unresolved support encounter in Student mode, allowing alternative tutoring responses to be explored and stored as reusable strategy memory.

## 4 Conclusion

RPO-PDT demonstrates how participant verification can be coupled with explicit conversational-role management in an embodied student-support agent. Its contribution lies in integrating confidence-controlled role transitions, conservative fallback behaviour, privacy-aware enrolment and retention, manual override, and modular dialogue control. The demonstration provides a concrete platform for examining social-role legibility, trust, privacy, and appropriate autonomy in participant-aware conversational agents.